# TOWARD A CATEGORY THEORY DESIGN OF ONTOLOGICAL KNOWLEDGE BASES

Nikolaj Glazunov
*National Aviation University, pr. Komarova 1, Kiev, Ukraine*
*glanm@yahoo.com*

Keywords: problem solving; knowledge base; representation theory; ontology; category theory; computer supported education.

Abstract: I discuss ontologies_and_ontological_knowledge_bases/formal_methods_and_theories duality and its category theory extensions as a step toward a solution to Knowledge-Based Systems Theory. In particular I focus on the example of the design of elements of ontologies and ontological knowledge bases of next three electronic courses: Foundations of Research Activities, Virtual Modeling of Complex Systems and Introduction to String Theory.

## 1 INTRODUCTION

The subject matter of this paper lies in the area between problem solving and formal methods (here by formal methods we will understand algebraic and category theory methods).

More specifically, I want to discuss the relations between the problems of design of ontologies and ontological knowledge bases of electronic courses and formal methods, theories and software for supporting this design and solving the problems and tasks of education.

The last start with the problem domain analysis, continues via requirements identification and specification. I will try to show that these two topics
are intimately related, and that the study of each is of the great importance for the development of the other.

The paper deals with formal knowledge-oriented methods. These include methods to development: (i) the structure of a course; (ii) the content of the course; (iii) computer and knowledge base support systems (Semantic Web and knowledge base Ontology are included).

A number of formal knowledge-oriented methods are presented in the literature: formal concept analysis (Ganter, Bernhard; Wille, Rudolf, 1998), (Ganter, Bernhard; Stumme, Gerd; Wille, Rudolf, eds., 2005), concept maps (Joseph D. Novak & Alberto J. Canas, 2006), conceptual graphs (Sowa, John F., 1976), semantic networks (Steyvers, M.; Tenenbaum, J.B., 2005).

In the paper we present the category theory interpretation some of these results, give ontologies_and_ontological_knowledge_bases / formal_methods_and_theories duality and its category theory extensions as a step toward a solution to Knowledge-Based Systems Theory.

We also illustrate our approach on the example of the design of elements of ontologies and ontological knowledge bases of next three electronic courses: Foundations of Research Activities, Virtual Modeling of Complex Systems and Introduction to String Theory.

## 2  COURCES, KNOWLEDGE, CONTENTS

Classical courses are stable and based on books and lectures. The novel courses are dynamical. They have many sources that include along with classical sources Internet and tutorial systems.

There are various states of knowledge that result from acquisition and derivation. Knowledge production involves stocks and flows of knowledge. The flows are concerned with acquiring knowledge and deriving knowledge. It is possible to associate acquiring knowledge with data/knowledge base and deriving knowledge with production, modelling, computer algebra and validated numeric systems. Both are typically interspersed in a decision-making process as for instance a justification of a conjecture and there can be tradeoffs between acquiring and deriving needed knowledge. There is a remarkable convergence of the problems of knowledge inference, dynamical control and controlled systems with the methods of homological algebra and category theory.

We shortly treat the topic of knowledge management. As a field of study, it is concerned with the representation and processing of knowledge. Here, we focused on knowledge management from angles of knowledge bases, homological algebra, computer algebra and validated numeric.

We apply and use formal methods at several numbers of levels. In general the application is not a linear process but some graph. Here by a graph we understand a collection of points and lines connecting some (possible empty) subset of them. We will consider graphs of different sorts (simple graphs, multigraphs, pseudographs). The edges, vertices, or both of a graph may be assigned specific values, labels, marks or colours. The edges of graphs may also be imbued with directedness. with m quiver. Not all aspects of the problem solving are formal. But it seems that it is possible to represent the description of the problem solving system as an interconnected collection of categories, and the progressive refinement of such descriptions from the requirements level to executable code (Goguen, 1989; Cadish & Diskin, 1995). In the case all aspects of system development are explicit and formal: the categories themselves, relationships between categories (functors/arrows), interconnections of categories (diagrams), refinements between categories and design records.

In the paper we illustrate our approach on the example of the design of elements of ontologies and ontological knowledge bases of next three courses: Foundations of Research Activities, Virtual Modeling of Complex Systems and Introduction to String Theory. These three courses are not independent. Foundations of Research Activities is a course with the problem domain "Electromagnetism and Gravitation" that enables the student to both apply and expand previous content knowledge toward the endeavour of engaging in an open-ended, student-centered investigation in the pursuit of an answer to a question or problem of interest. Some background in general relativity and quantum field theory are presented. The particular feature of the course is studying and applying computer-assisted methods and technologies to justification of conjectures (hypotheses). In our course, justification of conjectures encompasses those tasks that include gathering and analysis of data, go into testing conjectures, taking account of mathematical and computer-assisted methods of mathematical proof of the conjecture. Justification of conjectures is critical to the success of the solution of a problem. Design involves problem-solving and creativity. Virtual Modeling of Complex Systems is a course that study processes of problem-solving and planning for a virtual reality solutions. The main goal of the course "Introduction to String Theory" is to give a pedagogical introduction to string theory follow to the framework of investigations of the problem: Does string theory provide physics with a grand unification theory? It is primary intended for graduate students who are interested in string theory or quantum gravity. This course should be accessible for any listener with some background in general relativity and quantum field theory.

Below we will illustrate our approach by examples of the design of elements of ontologies and ontological knowledge bases for mentioned three courses.

## 3 ONTOLOGIES AND UPPER ONTOLOGIES

There are several definitions of the notion of ontology. By T. R. Gruber (Gruber, 1992) "An ontology is a specification of a conceptualization". By B. Smith and his colleagues, (Smith, 2004) "an ontology is a representational artefact whose representational units are intended to designate universals in reality and the relations between them". By our opinion the definitions reflect critical goals of ontologies in computer science. For our purposes we will use more specific definition of ontology: concepts with relations and rules define ontology (Gruber, 1992; Ontology, 2008; Wikipedia, 2009 ).

There are next relations among concepts:

associative
partial order
higher
subordinate
subsumption relation (is a, is subtype of, is subclass of)
part-of relation.

Different spaces are used in aforementioned courses. In our framework we treat ontology of spaces and ontology of symmetries as upper ontologies.

### 3.1 Elements of the ontology of spaces and symmetries

There is the well known space/ring_of_functions_on_the_space duality. In the subsection we only mention some concepts, relations and rules of the ontology of spaces and symmetries.

Two main concepts are space and symmetry.

3-dimensional real space $R^3$; Linear group $GL(3, R)$ of automorphisms of $R^3$;

Classical physical world has three spatial dimensions, so electric and magnetic fields are 3-component vectors defined at every point of space.

Minkowski space-time $M^{1,3}$ is a 4-dimensional real manifold with a pseudoriemannian metric $t^2 - x^2 - y^2 - z^2$. From $M^{1,3}$ it is possible to pass to $R^4$ by means of the substitution $t \to iu$ and an overall sign-change in the metric. A compactification of $R^4$ by means of a stereographic projection gives $S^4$. Let $SO(1,3)$ be the pseudoortogonal group. The moving frame in $M^{1,3}$ is a section of the trivial bundle $M^{1,3}$ $SO(1,3)$. A complex vector bundle $M^{1,3}$ $C^2$ is associated with the frame bundle by the representation $SL(2,C)$ of the Lorentz group $SO(1,3)$.

The space-time $M$ in which strings are propagating must have many dimensions (10, 26. …) . The ten-dimensional space-time is locally a product $M = M^{1,3}$ $K$ of macroscopic four-dimensional space-time and a compact six-dimensional Calabi-Yau manifold $K$ whose size is on the order of the Planck length.

Principle bundle over space-time, structure group, associated vector bundle, connection, connection one-form, curvature, curvature form, norm of the curvature.

2D space, 2D object, 3D space, 3D object.

Foregoing concepts with relations and rules define elements of the domain ontology of spaces.

### 3.2 Ontology for the visualization of a lecture

Upper ontology: visualization. Visualization of the text (white text against the dark background) is subclass of visualization.

Visible page, data visualization, flow visualization, image visualization, spatial visualization, surface rendering, two-dimensional field visualization, three-dimensional field visualization, video content.

## 4   PROBLEM ONTOLOGIES

Scientific Method.
Investigative processes, which are assumed to operate iteratively, involved in the research method are the follows:
(i)      Hypothesis, Low, Assumption, Generalization;
(ii)     Deduction;
(iii)    Observation, Confirmation;
(iv)    Induction.
   Indicate some related concepts: Problem. Class of Scientific Data. Scientific Theory. Formalization. Interpretation. Analyzing and Studying of Classic Scientific Problems. Investigation. Fundamental (pure) Research. Formulation of a Working Hypothesis to Guide Research. Developing Procedures to Testing a Hypothesis. Analysis of Data. Evaluation of Data.

### 4.1   Concepts of Foundations of Research Activities

Foundations of Research Activities Ontology includes upper and problem ontologies.
Foundations of Research Activities Concepts:
  (a) Scientific Method.
  (b) Ethics of Research Activity.
  (c) Embedded Technology and Engineering.
 (d) Communication of Results (Dublin Core).

   We consider briefly (a) and planning to consider briefly (c).

## 5   DOMAIN ONTOLOGIES

### 5.1  Concepts of Electrodynamics and Classical Gauge Theory

Short history: Schwarzchild action, Hermann Weyl, F. London, Yang-Mills equations.

Quantum Electrodynamics is regarded as physical gauge theory. The set of possible gauge transformations of the entire configuration of a given gauge theory also forms a group, the gauge group of the theory. An element of the gauge group can be parameterized by a smoothly varying function from the points of space-time to the (finite-dimensional) Lie group, whose value at each point represents the action of the gauge transformation on the fiber over that point.

Concepts: Gauge group as a (possibly trivial) principle bundle over space-time, gauge, classical field, gauge potential.

### 5.2 Concepts of another ontologies

Some Virtual Modeling of Complex Systems concepts:
Virtual;
Virtual Model, digital description of the object, the classical example of virtual model;
Virtual function, virtual class.
Modeling;
System;
Complexity.
The model and design of a virtual school are given by (Glazunov and Borovik, 2009).

Some ontologies of the course "Introduction to String Theory" have presented in (Glazunov, 2008a, 2008b)

## 6 FORMALIZATION OF ONTOLOGY

The formalization of steps (i) - (iv) from (a) carry out our considerations to
(v)     Axioms, Postulates;
(vi)    Rules of Inference;
(vii)   Confirmation (Verification);
(viii)  Induction (Uncertainty, Statistics, Probabilistic Rules).

Gruber and Olsen (Gruber, 1993) gave the ontology for physical quantities, units of measure and algebra for engineering models. We will use it for presentation of elements of ontology of charges and fields from the problem domain Electromagnetism.
The SI unit of quantity of electric charge is the coulomb. The charge of an electron is approximately $-1.602 \times 10^{-19}$ COULOMB.

(defobject COULOMB  (basic-unit COULOMB))

(defobject   NEWTON (basic-unit   NEWTON))

The electric field is a vector field with SI units of newtons per coulomb (NEWTON COULOMB$^{-1}$).

(defobject   NEWTON / COULOMB
     (= NEWTON / COULOMB
       (unit* newton (unit^   coulomb -1))))

Next step of the formalization is given by the following postulate:
Ontology as well as investigative method (that also may be treated as an ontology) may be represented as graphs (simple graphs, multigraphs, pseudographs).
We will call the graph the ontology graph.

## 7 ONTOLOGIES AND CATEGORY THEORY

### 7.1 Graphs and Categories

Recall definitions and basic facts about graphs and categories. Let ***Ens*** be the category of sets, $V, E$ its objects. A graph $G = (V, E, s, t)$ consists of sets $V, E$ and pair of functions $s, t: E \rightarrow V$. The functions $s$ and $t$ are called source and target and elements of $V$ and $E$ are called vertices and arrows. A morphism

$$f : G_1 \rightarrow G_2 \qquad (1)$$

of graphs is a pair of functions

$$f_v: V(G_1) \to V(G_2) \quad (2)$$

$$f_e: E(G_1) \to E(G_2) \quad (3)$$

such that $s(f_e(e)) = f_v(s(e))$, $t(f_e(e)) = f_v(t(e))$. The category **Graphs** consists on graphs and morphisms of graphs.

Let **Cat** be the category of small categories, **C** a small category. Recall two known results: the forgetful functor

$$F : \textbf{Cat} \to \textbf{Graphs}$$

assigns the underlying graph

$$G = G(C) = (V, E, s, t)$$

to the small category **C**, where $V(C)$ is the set of arrows, $E(C)$ of arrows of **C**. The left adjoint

$$P: \textbf{Graphs} \to \textbf{Cat}$$

of *F* assigns the free category **PathG** to a graph *G*. The category **PathG** is the category of paths of the graph *G*. The category **PathG** defines new graph

$$(V, PathG, s, t).$$

The last graph is used under construction of the productions system.

### 7.2 Categorification and Representation

If the ontology graph is finite, directed and connected (with possibly loops and multiple arrows) we will say that the corresponding ontology or method accept a *categorification*.

The graphs are called *quivers*. In many cases the graph is a directed tree.

There are several levels of algebraization and categorification of ontologies.

Let *Q* be a quiver without oriented cycles with $V = \{1,..,n\}$ the set of vertices and *E* the set of edges. Let *k* be a field. Let $i, j \in V$, $a \in E$. A representation of *Q* is a family

$$M = ((M(i)), M(a)) \quad (4)$$

where $M(i)$ is a finite dimensional $k$ – vector space and for every arrow $a: i \to j$ in *Q*,

$$M(a): M(i) \to M(j) \quad (5)$$

is a $k$ – linear transformation. Morphisms between representations are defined by the common way. This Gabriel (Gabriel, Roiter 1992) construction defines the category of representations $rep_k(Q)$.

Indicate some related notions.

Let $A$ be an algebra over a field $k$ and $V$ a vector space. A representation (Drozd, Kirichenko (1980)) of $k$-algebra $A$ is a homomorphism $h$ from $A$

$$h: A \to M(A), \quad (6)$$

to the algebra $M(A)$ of linear operators of $V$. If $h$ is an isomorphism

$$h(A) \approx A \quad (7)$$

then the representation is the exact.

Given a quiver $Q$ its path category **PathQ** has as objects the vertices of $Q$ and the morphisms

$$i \to j \quad (8)$$

are the paths from $i$ to $j$ which are by definition the formal compositions

$$a_1, \ldots, a_n \quad (9)$$

$a_n$ starts in $i$, $a_1$ ends in $j$ and the end point of $a_{i+1}$ coincides with the start point of $a_i$ for all

$$i = 1, \ldots, n-1. \quad (10)$$

We extend **PathQ** to the linear category **PathLQ** with the same class of objects and with morphisms *Mor (i, j)* = {the set of all paths from $i$ to $j$ and all their linear combinations with coefficients from $k$}.

An equivalent definition of the representation of a quiver is an additive functor from **PathLQ** to the category of $k$-**mod.**

A quiver $Q$ is marked (Roiter, (2001)) if for each $i, j \in V$ is attached a category $C_i$, and to each arrow $a: i \to j$ in $Q$ is attached a functor

$$\underline{F_a}: (C_i)^0 \times C_j \to \mathbf{Ens} \quad (11)$$

taking values in the category of sets.

A *categorification* is the category theory representation of the world that we want to investigate.

If an ontology has a representation by a category then we may apply homological and homotopical algebra constructions to study the ontology (Glazunov, 2007, 2006).

Below we give category theory concepts for our framework (an category theory ontology is given by Robert Kent (Kent, 2002)).

Category $C$ (small or big), object ($X, Y, \ldots$ ), morphism, class of morphisms $Hom_C (X, Y)$ ($C(X, Y)$ or $Mor_C(X, Y)$) or set of morphisms $hom_C (X, Y)$ ($C(X, Y)$ or $mor_C(X, Y)$) of $C$, dual category, monoid (a category with a single object $x$ and with $\#hom(x, x) \geq 1$). Subcategory. Full subcategory. Category of binary relations. Let **Ens** be the category of sets, $\mathbf{Ann}_k$ be the category of commutative rings with unit over a commutative ring $k$. *Spec R*. The category **Grp** of all groups with group homomorphisms as morphisms. The category **Ab** of abelian groups and group homomorphisms.

Bicategory. A monoidal category (or tensor category) is a bicategory with one object. *n*-categories. Groupoid. Picard groupoid. Stack.

Let *Cat* be the 2-category of small categories. Functors $\underline{F}: C \to D$ can be thought of as morphisms of *Cat*. Full and faithfull embedding of categories

| | |
|---|---|
| $\forall X, Y \in Ob\ C$ | (12) |
| $Hom_C\ (X, Y) \to Hom_D\ (\underline{F}X, \underline{F}Y)$ | (13) |

Natural transformation. Covariant and contravariant functors. Forgetfull functor. Representable functors. Pre-Sheaves and Sheave

| | |
|---|---|
| $(X, O_X)\ \ U \subset X\ ,\ O_X(U)$ | (14) |

Categories of functors or Functor categories (morphisms in this category are natural transformations between functors), $Ann_kE$, $Ann_kE\ (\underline{F},\underline{G})$. *K*-functor. *K*-tensor functor.

## 8  ONTOLOGICAL KNOWLEDGE BASES

By ontological knowledge base (KB) we will understand the KB that consist of in some way organized ontologies. In the full version of the paper we will present elements of the KB.

E-Learning lessons of the course "Foundations of Research Activities" are designed to guide students through information about the history and facts related to a research activities in the problem domain "Electromagnetism and Gravitation" and to help students perform specific tasks in the framework of computer-assisted methods and technologies to justification of conjectures (hypotheses). The cource is based on books and paper by Feynman R, Leighton R, and Sands M. (Feynman, Leighton, Sands, 1989), Landau L. D. and Lifshitz E. M. (Landau, Lifshitz, 1995), Hawking S., Ellis G., Landshoff P., Nelson D., Sciama D., Weinberg S ( Hawking, Ellis, Landshoff, Nelson, Sciama, Weinberg, 1975) and on Wikipedia (Wikipedia, 2009). E-Learning lessons of the course "Virtual Modeling of Complex Systems" are designed to guide students through information about the history and facts related to virtual modeling of complex systems in the problem domain "Virtual Reality" and to help students perform specific tasks under designing elements of virtual reality (elements of virtual classrooms are included). E-Learning lessons of the course "Introduction to String Theory" are designed to guide students through information about the history and facts related to elements of closed and open strings (Polchinski, 1998) (Wess and Bagger, 1983) (Glazunov, 2008a, 2008b) with applications to the problem domain "Electromagnetism and Gravitation" and to help students perform specific tasks in the framework of closed and open strings.

## 9  DISCUSSION AND CONCLUSIONS

In the paper, we have discussed applications of formal methods, algebra and category theory to design of elements of ontologies and ontological knowledge bases of electronic courses. We have illustrated our approach on the example of the design of elements of ontologies and ontological knowledge bases of next three courses: Foundations of Research Activities, Virtual Modeling of Complex Systems and Introduction to StringTheory.

The next step towards a Knowledge Based Systems Theory of e-learning is a more complete understanding of the category theory basics of problem_solving/formal_methods dualities. There are several ways that on might try to approach this. I think that a promising one is to implement the categorification of space of problem solving (CSPS) and functions on the space (CFSPS) and to use CSPS/ CFSPS duality backwards.

# REFERENCES

Ganter, B., Wille, Rudolf, 1998. *Formal Concept Analysis: Mathematical Foundations*, Springer-Verlag, Berlin.

Ganter, B., Stumme, G., Wille, R., eds., 2005. *Formal Concept Analysis: Foundations and Applications*, Lecture Notes in Artificial Intelligence, no. 3626, Springer-Verlag.

Novak, J,. Canas, A., 2006. The Theory Underlying Concept Maps and How To Construct and Use Them, Institute for Human and Machine Cognition. Accessed 24 Nov 2008.

Sowa, J., 1976, Conceptual Graphs for a Data Base Interface, *IBM Journal of Research and Development* 20(4), 336–357.

Steyvers, M.; Tenenbaum, J.B., 2005. The Large-Scale Structure of Semantic Networks: Statistical Analyses and a Model of Semantic Growth, *Cognitive Science* **29** (1), 41–78.

Gruber, Thomas R. 1992. A Translation Approach to Portable Ontology Specifications. *Knowledge Systems Laboratory September 1992 Technical Report KSL 92-71.*

Goguen, J.A. 1989. A categorical manifesto. Technical report. *SRI International, SRI-CSL-89-8.*

Cadish, B., Diskin, Z. 1995. Algebraic Graph-Oriented = Category-Theory-Based. *NGITS'95 Workshop.*

Ontology. 2008, *Encyclopedia of Database Systems*, Ling Liu and M. Tamer Ozsu (Eds.), Springer-Verlag, 2008.

Smith, B. 2004. *Normalizing Medical Ontologies using Basic Formal Ontology*, in *Kooperative Versorgung, Vernetzte Forschung, Ubiquitare Information* (Proceedings of GMDS Innsbruck, 26—30 September 2004), Niebull: Videel OHG, 199—201.

Spear, A. 2006. *Ontology for the Twenty First Century: An Introduction with Recommendations*, IFOMI, Saarbrucken, Germany.

Wikipedia, 2009, the free encyclopedia.

Gruber, Thomas R., 1993. Toward Principles for the Design of Ontologies Used for Knowledge Sharing. In. *Technical Report KSL 93-04, Knowledge Systems Laboratory,* Stanford University.

Feynman R, Leighton R, and Sands M., 1989. *The Feynman Lectures on Physics*, 3 volumes.

Landau L. D., Lifshitz E. M. 1995. *Mechanics and Electrodynamics*. Hardcover, Elsevier Science & Technology.

Hawking S., Ellis G., Landshoff P., Nelson D., Sciama D., Weinberg S. 1975. *The Large Scale Structure of Space-Time,* Cambridge Monographs on Mathematical Physics.

Gabriel, P., Roiter, A. 1992. *Representations of Finite-Dimensional Algebras*, Springer-Verlag.

Drozd Yu.A., V.V. Kirichenko V.V., 1980. *Finite Dimensional Algebras*, Vyscha shkola, Kiev.

Roiter A.V., Representations of marked quivers, E-print http://arXiv.org/abs/math.RT/0112158.

Kent Robert E., 2002. The IFF Category Theory Ontology.

Polchinski, J., 1998. *String theory* I, Cambridge Univ. Press, Cambridge.

Wess J., Bagger J., 1983. *Supersymmetry and supergravity,* Princeton Univ. Press.

Glazunov, N., 2008a. Homological and Homotopical Algebra of Supersymmetries and Integrability to String Theory, *Proceedings of Int. Workshop "New Trends in Science and Technology" (NTST08)*). http://ntst08.cankaya.edu.tr/proceedings/

Glazunov, N., 2008b. Homological and Homotopical Algebra of Supersymmetries and Integrability to String Theory (introduction and preliminaries). *E-print. arXiv: 0805.4161*

Glazunov, N., 2007. Object-Oriented Knowledge Base System for Reasoning About OWL-Ontologies. *Final Scientific Report*. Kiev. 2007.

Glazunov, N., 2007. Semantic Web and Category Theory Models. *Proc. Int. Seminar "Discrete Mathematics and its Applications"*. Moscow: MGU (2007) P.309-311.

Glazunov, N., 2006. Category Theory Models of Discrete Control Systems. *Seventh International Conference "Discrete Models in the Theory of Control Systems*, Moscow Lomonosov State University, Russia, 4-6 March 2006.

Glazunov, N., Borovik, V. 2009. *Databases in document-information sphere*, National Aviation Univ., Kiev.